\documentclass[pdflatex,sn-basic,Numbered]{sn-jnl}

\usepackage{graphicx}%
\usepackage{multirow}%
\usepackage{amsmath,amssymb,amsfonts}%
\usepackage{amsthm}%
\usepackage{mathrsfs}%
\usepackage[title]{appendix}%
\usepackage{xcolor}%
\usepackage{textcomp}%
\usepackage{manyfoot}%
\usepackage{booktabs}%
\usepackage{algorithm}%
\usepackage{algorithmicx}%
\usepackage{algpseudocode}%
\usepackage{listings}%
\theoremstyle{thmstyleone}%

\usepackage{makecell}  
\usepackage{subcaption}
\theoremstyle{thmstyletwo}%
\theoremstyle{thmstylethree}%

\begin{document}

\title[Article Title]{SE-MLP Model for Predicting Prior Acceleration Features in Penetration Signals}

\author[1]{\fnm{Yankang} \sur{Li}}\email{lyk1520791162@njust.edu.cn}
\author*[1]{\fnm{Changsheng} \sur{Li}}\email{lichangsheng1984@163.com}

\affil*[1]{\orgdiv{School of Mechanical Engineering}, \orgname{Nanjing University of Science and Technology}, \orgaddress{\city{Nanjing}, \postcode{210094}, \state{Jiangsu}, \country{China}}}


\abstract{Accurate identification of the penetration process relies heavily on prior feature values of penetration acceleration. However, these feature values are typically obtained through long simulation cycles and expensive computations. To overcome this limitation, this paper proposes a multi-layer Perceptron architecture, termed squeeze and excitation multi-layer perceptron (SE-MLP), which integrates a channel attention mechanism with residual connections to enable rapid prediction of acceleration feature values. Using physical parameters under different working conditions as inputs, the model outputs layer-wise acceleration features, thereby establishing a nonlinear mapping between physical parameters and penetration characteristics. Comparative experiments against conventional MLP, XGBoost, and Transformer models demonstrate that SE-MLP achieves superior prediction accuracy, generalization, and stability. Ablation studies further confirm that both the channel attention module and residual structure contribute significantly to performance gains. Numerical simulations and range recovery tests show that the discrepancies between predicted and measured acceleration peaks and pulse widths remain within acceptable engineering tolerances. These results validate the feasibility and engineering applicability of the proposed method and provide a practical basis for rapidly generating prior feature values for penetration fuzes.}



\maketitle

\section{Introduction}\label{sec1}

Modern high-value targets, including critical military facilities and command centers, are frequently concealed underground or protected by multilayer protective systems composed of concrete, rock, and other high-strength media. Defeating such targets requires hard-penetrator warheads to withstand extreme impact and complex transient loads while acquiring target-related state information in real time. Reliable state awareness during penetration enables accurate identification of structural layers and penetration position, which is essential for precise fuze detonation control.

Most existing burst point control strategies for penetration fuzes~\cite{Liu2023,li2024lr,weng2024enhancing} fundamentally depend on \emph{prior feature values}. These values are typical dynamic-response characteristics that can be estimated before deployment from known physical conditions, such as warhead configuration, target parameters, and impact velocity. Across widely used decision logics, including threshold-based methods, envelope-based methods, and time-window shielding methods, the control rules are driven by comparisons between real-time sensor measurements and these pre-estimated priors. Representative priors include the acceleration peak, pulse width, interlayer time interval, penetration-energy-related characteristics, and dominant frequency components. In engineering practice, the threshold method remains the most common for penetration state identification. The onboard system continuously measures acceleration and triggers state transitions (entry, exit, or interlayer flight) when extracted features exceed or fall below preset thresholds. Accordingly, signal-processing modules primarily serve to make real-time features more comparable to the priors. This implies a critical dependency: even with sophisticated signal processing, inaccurate or unreliable prior feature values can still lead to layer miscounting, missed layers, and degraded burst point precision~\cite{Liu2023, Zhang2022}. Therefore, prior feature values are not only auxiliary parameters but a central determinant of recognition robustness and control accuracy.

In current practice, prior feature values of penetration acceleration are mainly obtained via empirical formulas, range recovery tests, and numerical simulations. These approaches, however, often entail long development cycles and high costs, and their outputs can exhibit notable dispersion and limited credibility, which in turn forces repeated tests or repeated simulation campaigns. Moreover, empirical methods~\cite{shen2024imagpose,shen2025imagdressing} tend to generalize poorly across diverse targets and operating conditions, while numerical simulations are sensitive to constitutive relations, contact settings, and model parameters. As a result, conventional prior acquisition pipelines struggle to support diversified missions, large-scale deployment, and rapid response. This motivates a practical alternative: a prediction method that directly maps physical condition parameters to critical acceleration prior features. Such a method can enable targeted offline computation prior to fuze-warhead deployment, and then embed the predicted priors into control software, improving engineering feasibility. In addition, if the predictor generalizes across working conditions, it can provide tunable and batch-generated prior features, establishing more reliable inputs for layer counting, state identification, and burst point control.

To reduce the burden of traditional physical modeling and improve efficiency, recent studies have introduced deep learning for penetration-dynamics prediction. A TCN-LSTM multisource fusion framework integrates acceleration and magnetic-anomaly signals for high-accuracy layer counting, and its outputs can be directly used for fuze-control decisions~\cite{WangY2024}. A POD-RBF model combines dimensionality reduction with radial basis regression to accelerate magnetic-response feature prediction under multiple conditions~\cite{WangYL2025}. For incorporating structural parameters, PF-Informer modifies the Transformer to fuse structural parameters with measured acceleration for real-time prediction of penetration acceleration histories~\cite{Ma2023}. TransUNet further combines Transformer and U-Net architectures to generate acceleration distributions across penetration depths and support data augmentation in multilayer scenarios~\cite{LiR2023}. While these methods advance prediction accuracy and information fusion, their relatively high model complexity, parameter scale, and computational cost hinder deployment on embedded platforms with strict resource constraints. Consequently, there remains a clear need for a predictor that is structurally simple, computationally efficient, and simultaneously accurate and deployable.

In this paper, we propose a lightweight predictor for penetration acceleration prior features, termed the squeeze and excitation multi-layer perceptron (SE-MLP). The SE-MLP takes physical parameters describing penetration conditions as inputs and outputs layer-specific acceleration feature values. The design enhances a standard MLP with squeeze-and-excitation channel reweighting and residual fusion to strengthen physical-parameter representation and stabilize optimization, enabling an efficient mapping from operating conditions to acceleration priors. To ensure objective evaluation and mitigate dataset bias, we adopt four-fold cross-validation ($K{=}4$) with strict separation between training and testing samples.

The main contributions are summarized as follows:

\begin{itemize}

  \item We formalize prior feature value prediction for penetration acceleration as a physics-parameter-to-feature mapping problem that directly supports burst point control and layer-state recognition.
  
  \item We propose SE-MLP, a lightweight MLP enhanced with squeeze-and-excitation attention and residual fusion, achieving an improved trade-off among accuracy, stability, and embedded deployability.
  
  \item We conduct rigorous evaluation using four-fold cross-validation with strict train-test separation to provide an objective assessment of generalization across penetration conditions.
  
\end{itemize}

\section{Model Architecture}\label{sec2}

In this section, a SE-MLP model integrating a multilayer perceptron structure, a channel attention mechanism, and residual connections is constructed to achieve a nonlinear mapping between physical characteristic parameters and the prior acceleration features. Considering that prediction models based on a single network structure generally suffer from insufficient generalization capability and limited robustness under complex penetration conditions, a channel attention mechanism is introduced to enable adaptive weighting of input features, while residual connections are incorporated to enhance feature preservation and gradient propagation.
The overall model consists of three components: a multi-layer perceptron (MLP) for building the fundamental mapping structure, a squeeze and excitation (SE) module for strengthening the adaptive weighting capability across feature channels, and residual connections for transmitting cross-layer information and optimizing gradient flow. The model takes as input the physical parameters under different penetration conditions from projectile penetration simulations (including projectile mass, impact velocity, target-layer number, and material parameters), and outputs the characteristic acceleration features. Through network learning, a mapping is established between the high-dimensional physical inputs and the nonlinear acceleration-response features. In this study, the acceleration peak and pulse width corresponding to each penetrated layer are selected as the prediction outputs to verify the feasibility of the proposed approach. Through training and validation under multiple operating conditions, the model can accurately predict the acceleration-response features corresponding to different target configurations, enabling efficient modeling of energy transfer and dynamic-response patterns during the penetration process. This method, to some extent, overcomes the limitations of traditional empirical formulas and shallow neural-network models, and provides a high-accuracy and robust machine-learning framework for penetration-dynamics feature prediction.

\subsection{Multi-layer Perceptron (MLP)}\label{subsec2}

A multi-layer perceptron (MLP) is a feed-forward neural network that establishes complex nonlinear relationships between inputs and outputs through a combination of multiple linear transformations and nonlinear activation functions. An MLP typically consists of an input layer, several hidden layers, and an output layer. As shown in Fig. 1, the output of each layer can be expressed as follows~\cite{Zeng2023,haykin2009neural,ChenSA2023,Dudek2020}:

\begin{equation}
    \label{eq:mlp_layer}
    \mathbf{h^{(l)}} = \mathrm{f}\left(\mathbf{W^{(l)}} \mathbf{h^{(l-1)}} + \mathbf{b^{(l)}}\right), \quad l = 1, 2, \ldots, L.
\end{equation}
Where $\mathbf{h^{(l)}}$ denotes the output feature vector of the $l$-th layer, while $\mathbf{W^{(l)}}$ and $\mathbf{b^{(l)}}$ represent the weight matrix and bias vector, respectively. The term $\mathrm{f}(\cdot)$ denotes the nonlinear activation function (e.g., GELU, ReLU). Through the stacking of multiple layers and nonlinear transformations, the MLP progressively abstracts features from the input data, thereby facilitating the learning of complex mappings in high-dimensional space.

However, with the increase of network depth, the MLP model is prone to gradient vanishing or gradient exploding problems during the training process, making the model convergence speed decrease and affecting the final performance. To alleviate this problem, the Residual Connection structure is introduced to mitigate the degradation problem of deep networks. Its core idea is to establish a cross-layer channel (Shortcut Path) for the network, enabling the input to be directly transmitted to the output end of subsequent layers, thereby guaranteeing the effective backpropagation of gradients. The residual structure can be expressed as~\cite{He2016}:
\begin{equation}
\mathrm{H}(x) = \mathrm{F}(x) + x.
\end{equation}
Wherein, $\mathrm{H}(x)$ represents the final output of the residual block, $x$ is the input feature, and $\mathrm{F}(x)$ represents the mapping function processed through several linear layers and nonlinear activation functions. Through this cross-layer stacking manner, the network can, while maintaining the integrity of the original information, learn the residual mapping between the input and the target, avoid feature loss during multi-layer transformations, and effectively improve the convergence speed and prediction accuracy of the model.

As shown in Fig. 2, the residual connection structure, through performing element-wise addition of the input $x$ and the output of the mapping function $\mathrm{F}(x)$, realizes the cross-layer transmission of features, can effectively alleviate the gradient vanishing problem, promote the training stability of the deep model, and enhance the fusion ability of low-level and high-level features.

\begin{figure}[t]
    \centering
    \begin{minipage}[b]{0.48\textwidth}
        \centering
        \includegraphics[height=5.5cm, keepaspectratio]{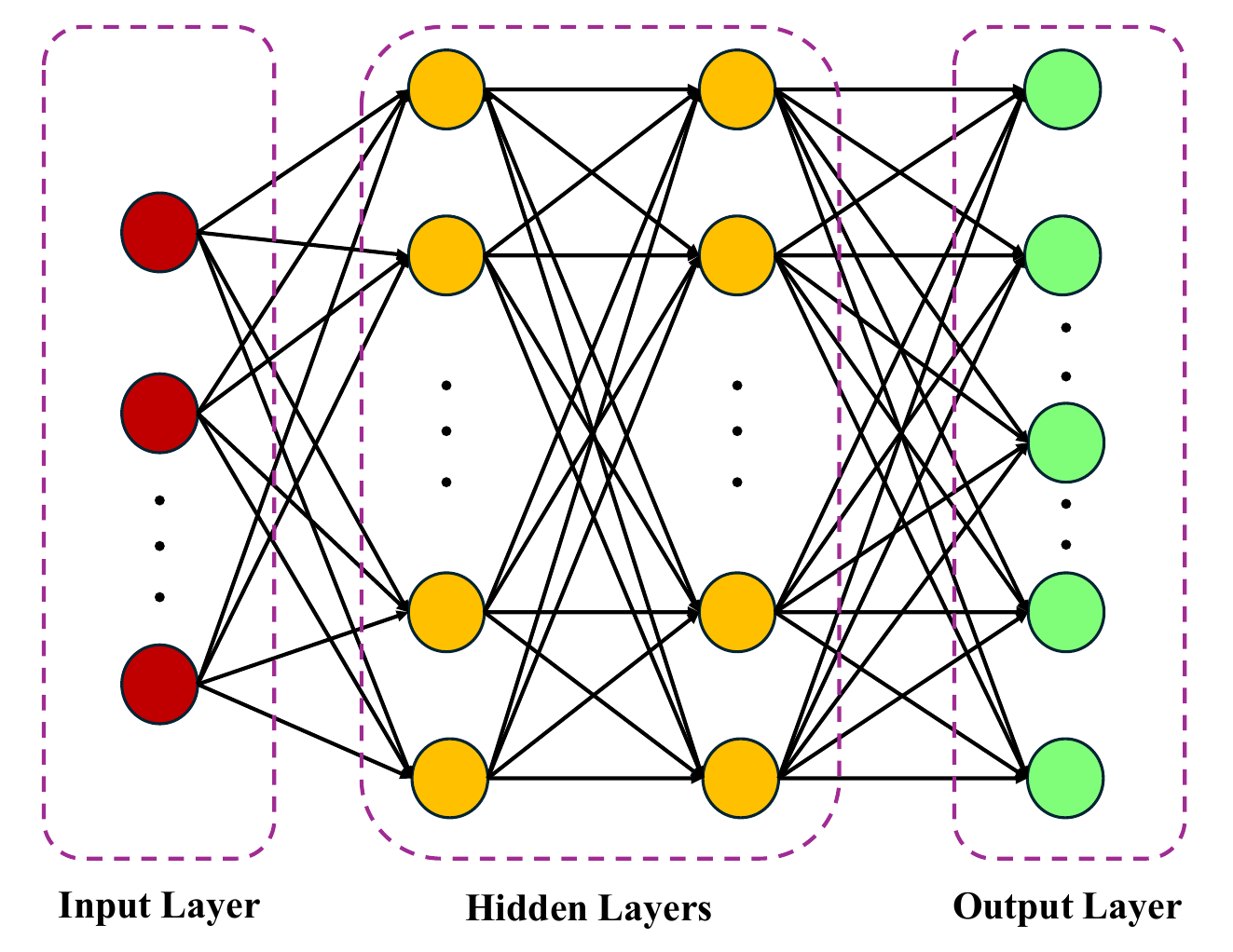}
        \caption{Schematic diagram of the MLP module.}
        \label{fig:mlp}
    \end{minipage}
    \hfill 
    \begin{minipage}[b]{0.48\textwidth}
        \centering
        \includegraphics[height=5.5cm, keepaspectratio]{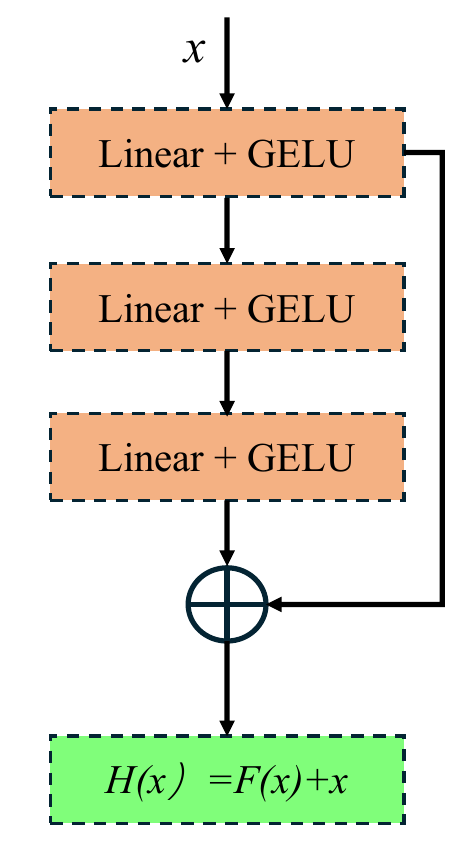}
        \caption{Schematic diagram of the residual module.}
        \label{fig:residual}
    \end{minipage}
\end{figure}

\subsection{Squeeze and Excitation Attention Mechanism (SE)}\label{subsec2}

The core idea of the squeeze and excitation (SE) attention mechanism is to, by explicitly modeling the dependencies between feature channels, realize the dynamic re-calibration of channel-level features, which can enhance the sensitivity of the model to key features. The SE module mainly includes two phases: Squeeze (compression) and Excitation (excitation), and its structural schematic is shown in Fig. 3~\cite{Hu2018}.

\begin{figure}[t]
    \centering
    \includegraphics[width=0.9\textwidth]{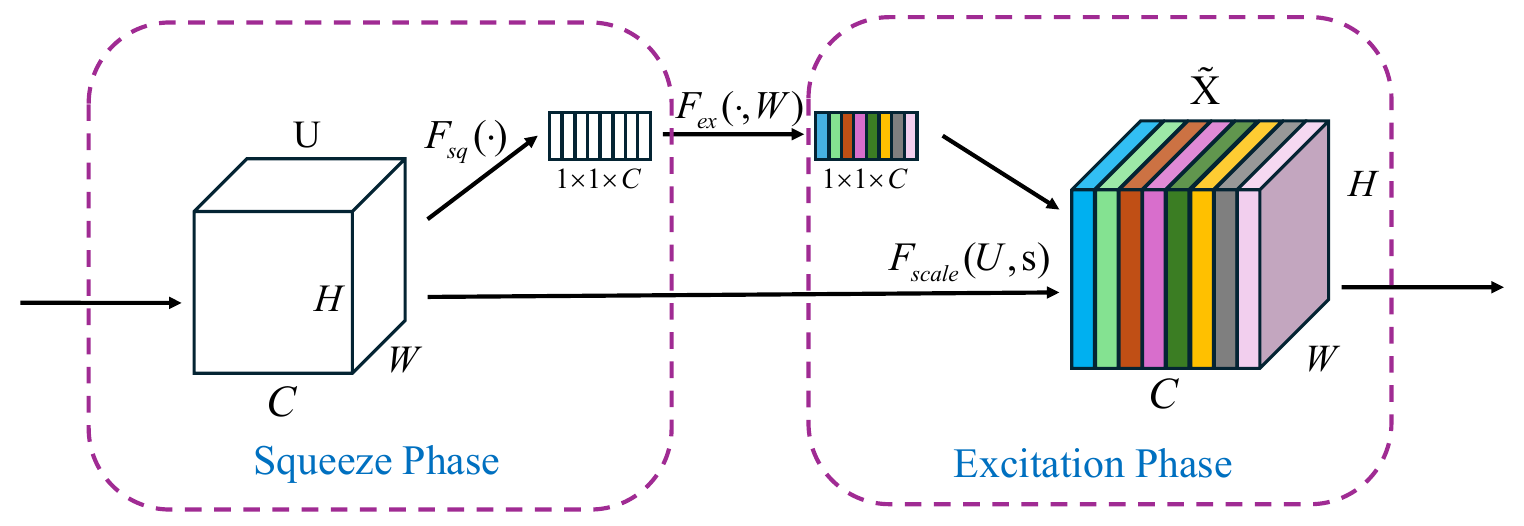}
    
    \caption{Structure diagram of the SE attention mechanism.}
    
    \label{fig:se_structure}
\end{figure}

\subsubsection{Squeeze phase (global information compression)}\label{subsubsec2}
In the Squeeze phase, by performing global average pooling on the input feature in the spatial dimension, the feature map $\mathbf{X} \in \mathbb{R}^{H \times W \times C}$ is compressed into the channel descriptor vector $\mathbf{z} \in \mathbb{R}^{C}$, and its calculation formula is:

\begin{equation}
    z_c = \frac{1}{H \times W} \sum_{i=1}^{H} \sum_{j=1}^{W} X_c(i, j).
\end{equation}
Wherein, $z_c$ represents the global feature response of the $c$-th channel, $H$ is the height, $W$ is the width, $c$ is the number of channels, and $X_c(i, j)$ represents the value of the $c$-th channel at position $(i, j)$. This process realizes the aggregation of global information, enabling the model to perceive the importance of each channel in the overall feature expression.

\subsubsection{Excitation phase (channel weight learning)}\label{subsubsec2}
In the Excitation phase, dependencies between channels are constructed through two fully connected layers and nonlinear activation functions, generating the channel weight vector $\mathbf{s}$:

\begin{equation}
    \mathbf{s} = \sigma\left(\mathbf{W_2} \cdot \delta\left(\mathbf{W_1} \cdot \mathbf{z}\right)\right).
\end{equation}
Wherein, $\delta(\cdot)$ represents the channel weight vector obtained through the GELU activation function, $\sigma(\cdot)$ denotes the Sigmoid function, and the weight matrices of the two fully connected layers $\mathbf{W_1}$ and $\mathbf{W_2}$, reflecting the importance degree of each channel; finally, feature enhancement is completed through the channel re-calibration $\mathrm{F}_{scale}(\cdot)$ operation:

\begin{equation}
    \tilde{X}_c = \mathrm{F}_{scale}(X_c, s_c) = s_c \cdot X_c.
\end{equation}
Wherein, $\tilde{X}_c$ is the feature of the $c$-th channel, and $s_c$ is the corresponding attention weight. The weighted feature $\tilde{X}_c$ will significantly enhance the response of key channels and suppress irrelevant information, thereby improving the representation ability and generalization performance of the model.

\subsection{Construction of the SE-MLP Model}\label{subsec2}
Building upon the aforementioned multi-layer Perceptron (MLP) and the squeeze and excitation (SE) attention mechanism, this paper integrates the strengths of both to construct a three-layer SE-MLP model based on channel attention and residual fusion. Taking the physical characteristic parameters from penetration numerical simulations as input, the model extracts feature representations hierarchically through the three-layer MLP structure, and introduces an SE module after each layer to adaptively adjust the weights of feature channels, thereby highlighting key feature dimensions and suppressing irrelevant or redundant information. Compared with traditional MLPs, the SE-MLP significantly improves the network's capability to automatically identify the importance of input features while maintaining a compact structure, thereby enhancing its nonlinear modeling capability and generalization performance. Furthermore, residual connection modules are incorporated into the model structure to mitigate the problems of gradient vanishing and feature degradation commonly found in deep networks through cross-layer feature fusion, enabling the network to maintain stable gradient flow and effective feature transmission when processing high-dimensional physical parameters. Finally, the network outputs the predicted penetration acceleration  characteristic values through a linear mapping layer and employs the Weighted Mean Squared Error (WMSE) as the loss function to emphasize the learning of key target features, achieving high-precision prediction of penetration characteristic values under multiple working conditions. The overall structure of the model is shown in Fig. 4.

\begin{figure}[t]
    \centering
    \includegraphics[width=0.9\textwidth]{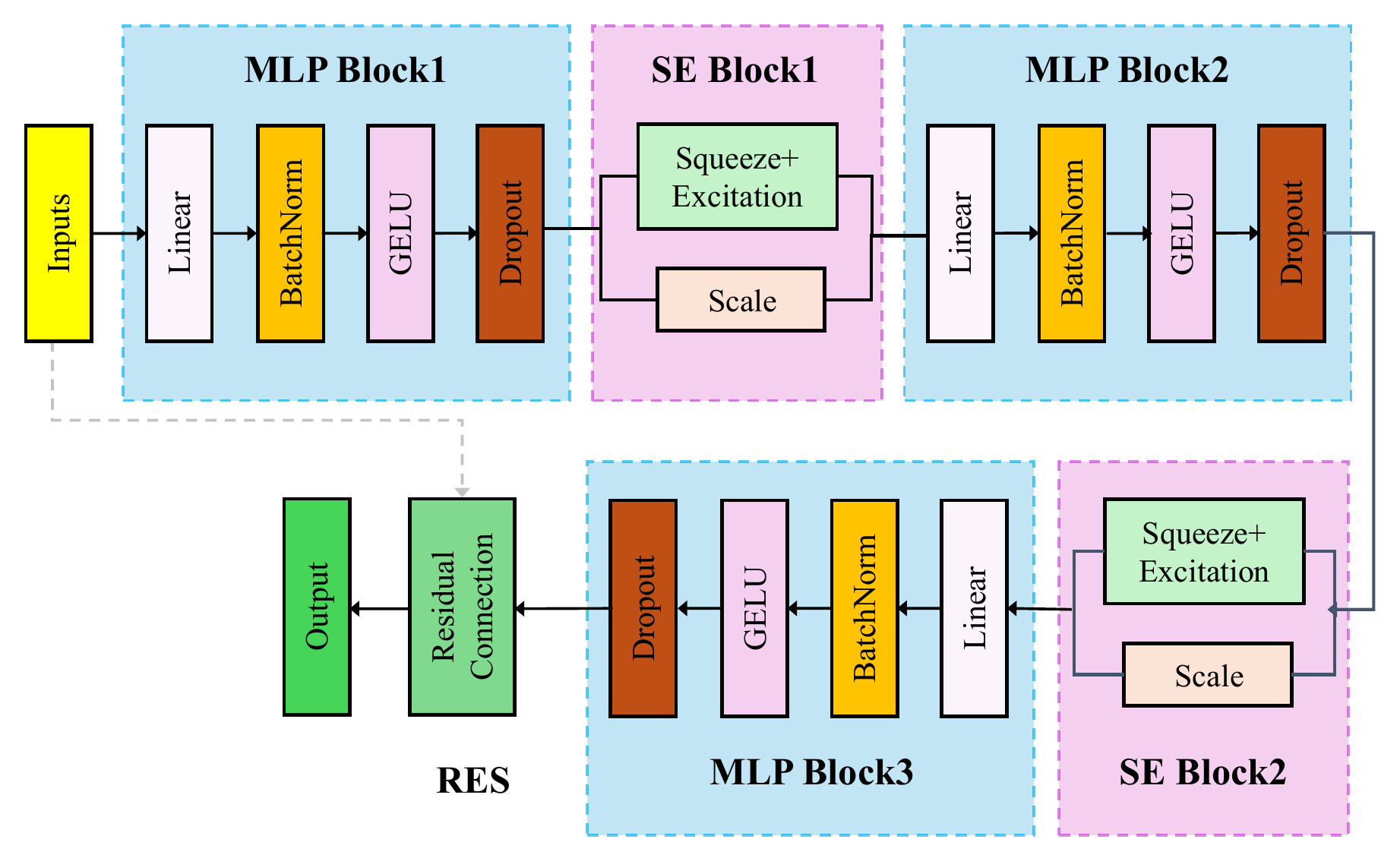}
    
    \caption{Structure diagram of the squeeze and excitation multi-layer perceptron (SE-MLP).}
    
    \label{fig:se_structure}
\end{figure}

The physical feature parameters of the warhead are taken as model inputs and fed into the network after unified scale processing. The main body of the model is composed of three-layer fully connected structures, where each layer contains a Linear fully connected layer, a Batch Normalization (BN) layer, a GELU activation function, and a Dropout layer to enhance training stability and reduce the risk of overfitting. An SE attention module is introduced after the output of each fully connected layer to learn the global importance weights of each channel through "squeeze" and "excitation" operations, achieving dynamic re-calibration of inter-channel features. This mechanism enables the model to automatically focus on key feature dimensions that contribute significantly to the prediction results, thereby significantly enhancing the feature representation capability and improving prediction accuracy.

After three layers of MLP feature extraction with attention, the model fuses the extracted features with the input signals through a residual connection module to realize cross-layer information transmission, effectively alleviating the gradient vanishing problem and further enhancing the training stability and generalization performance of the model. Finally, the acceleration signal feature values are output through a linear mapping layer, and this paper selects the acceleration peak and pulse width of each penetration layer as the target feature values for prediction. To ensure consistency between numerical scale and physical meaning, the acceleration peak adopts natural logarithm transformation for amplitude compression to alleviate the instability caused by its cross-scale variation to network training; the pulse width, as a low-amplitude time feature, adopts max normalization to maintain scale consistency among different samples.

\section{Data Acquisition of Penetration Fuze Acceleration Signals}\label{sec3}

The input of the model in this study consists of key physical feature parameters of the projectile-target system, primarily including the warhead type, impact velocity, target layer count, and material parameters, as well as the acceleration signal feature values for each layer obtained from numerical simulations. All input data are derived from numerical simulation results based on experimentally validated models, integrated with partial range test data. Through systematic modeling and parameter control, the coupling effects and complementary characteristics among physical parameters under various penetration conditions are fully considered. This approach effectively overcomes the limitations associated with using single-parameter inputs to characterize the penetration process.

In the process of dataset establishment, to ensure that the model possesses good generalization ability and data representativeness, typical penetration working conditions are selected for parameter combination to conduct numerical simulations, covering different warhead types, initial velocities, target materials, and penetration layers. Due to the complexity of the simulation calculation process, an increase in data volume will significantly escalate the computational cost; therefore, under the premise of guaranteeing a balance between accuracy and efficiency, this paper selects an appropriate sampling rate and dataset scale for simulation. Through multiple verifications, the sampling rate of the acceleration response signal is set to 40 kHz (i.e., the interval between adjacent sampling points is 0.025 ms) to ensure the capture of high-frequency impact features while accounting for computational resource efficiency. The multi-condition dataset obtained through the aforementioned methods is used as the model input after feature scale unification processing, providing high-quality sample support for the feature learning and generalization of the SE-MLP model.The simulation dataset is not publicly available due to confidentiality restrictions.

This paper selects different warheads penetrating multi-layer concrete targets of different materials as the research object and establishes a full-scale numerical simulation model to simulate a total of 108 working conditions; the specific simulation working condition parameters are shown in Table 1.

\begin{table}[t]
    \centering
    \small
    \caption{Parameters of simulation working conditions.}\label{tab:sim_params}
    
    \begin{tabular*}{\textwidth}{@{\extracolsep\fill}lc}
    
        \toprule
        Physical feature parameters & Penetration working conditions \\
        
        \midrule
        Warhead type & 60 kg-class, 150 kg-class, 300 kg-class, 600 kg-class \\
        Penetration velocity & 900 m/s--1700 m/s (step size is 100 m/s) \\
        Target material type & C40, C60, C80 \\
        Number of penetration layers & 6--10 \\
        
        \botrule 
    \end{tabular*}
\end{table}

\begin{figure}[t]
    \centering
    \includegraphics[width=0.5\textwidth]{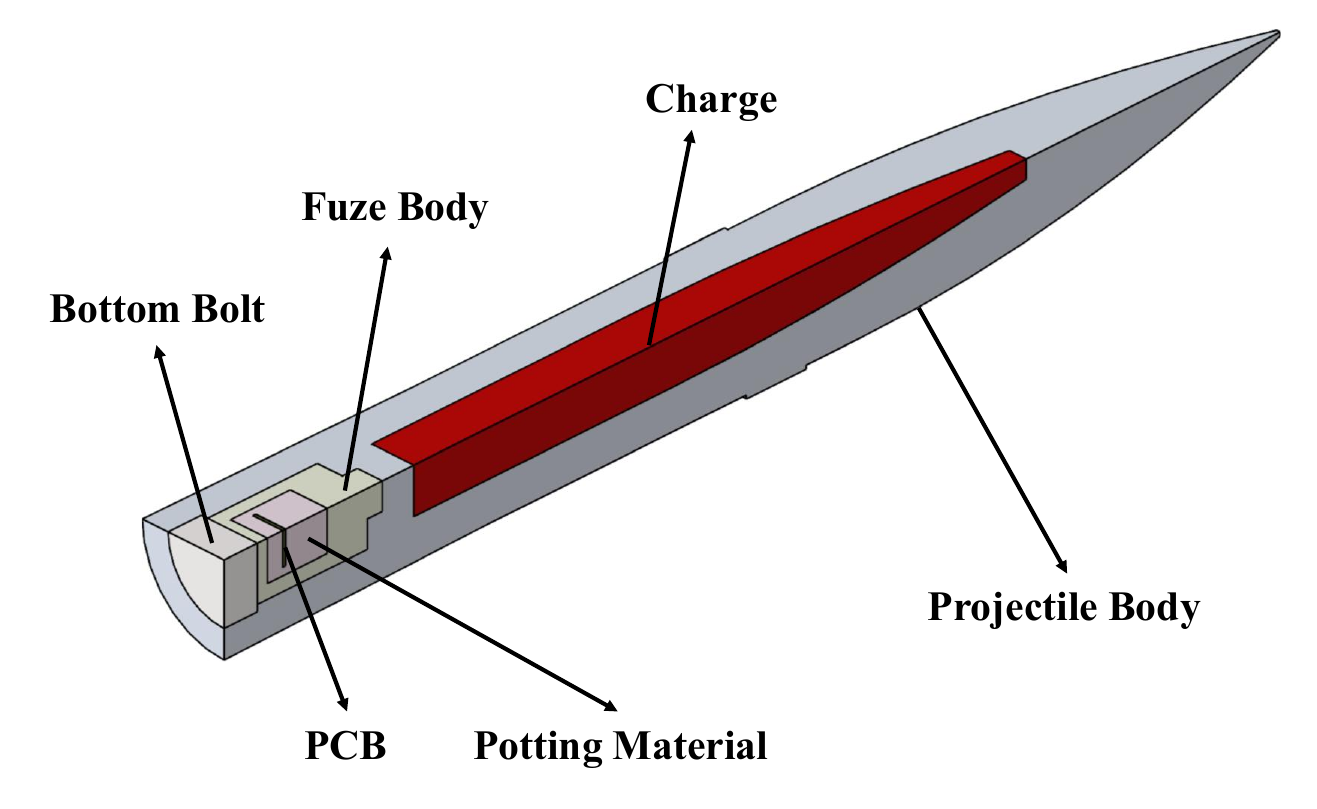}
    
    \caption{1/4 symmetry model of the typical structure of the projectile-fuze system.}  
    \label{fig:se_structure}
\end{figure}

\begin{figure}[t]
    \centering
    \includegraphics[width=0.8\textwidth]{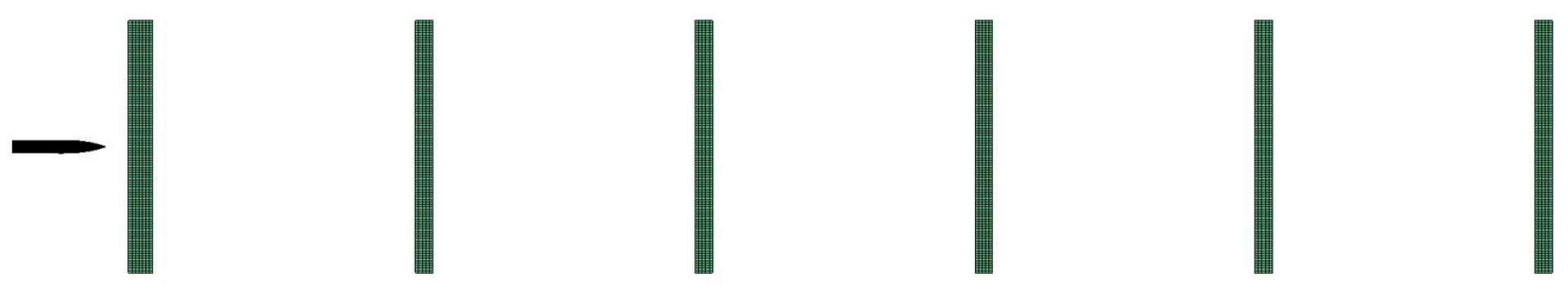}
    
    \caption{Numerical simulation model of projectile–target penetration.}
    
    \label{fig:se_structure}
\end{figure}

The warhead as a whole is composed of the projectile body, warhead main charge, fuze, base screw, and other parts, wherein the fuze is simplified into a shell, potting material, and a PCB, and a gasket is placed at the connection interface between the fuze and the projectile body to play the role of acceleration buffering. The projectile body and base screw both adopt high-strength alloy steel material, while the fuze shell adopts aluminum alloy material; based on this, four different projectile-fuze systems are established respectively according to the warhead parameters, and the 1/4 symmetry model of the typical structure of the projectile-fuze system is shown in Fig. 5~\cite{Peng2023}.

\begin{table}[t]
    \centering
    \caption{Simulation material parameters.}
    \label{tab:material_params}
    
    \setlength{\tabcolsep}{4pt}
    
    \begin{tabular*}{\textwidth}{@{\extracolsep{\fill}}lccccc}
        \toprule
        Material & 
        \makecell[c]{Material density \\ $\rho/(\mathrm{kg}/\mathrm{m}^3)$} & 
        \makecell[c]{Elastic modulus \\ $E/\mathrm{GPa}$} & 
        \makecell[c]{Yield stress \\ $\sigma/\mathrm{MPa}$} & 
        \makecell[c]{Tangent modulus \\ $E_{\mathrm{tan}}/\mathrm{GPa}$} & 
        \makecell[c]{Poisson's ratio \\ $\mu$} \\
        
        \midrule
        
        Projectile body & 7800 & 207 & 1720 & 2.1 & 0.3 \\
        Charge & 1760 & 20.1 & 200 & 0 & 0.35 \\
        Base screw & 7800 & 207 & 1720 & 2.1 & 0.3 \\
        Potting material & 580 & 24.1 & 350 & 0 & 0.3897 \\
        PCB & 1500 & 11 & 100 & 0 & 0.35 \\
        Fuze body & 2780 & 72.4 & 345 & 0.777 & 0.33 \\
        
        \bottomrule
    \end{tabular*}
\end{table}

The target is set as 6–10 layers of concrete targets, wherein the thickness of the first layer target is 0.3 m, and the thickness of each remaining layer target is 0.18 m, the material strength grade covers C40–C80, and the target spacing is 3 m. Fig. 6 shows the schematic of the projectile–target penetration simulation model.

The material parameters of the warhead and fuze are shown in the table 2~\cite{Xin2019}.

\section{Performance Verification of the Network Model}\label{sec4}

\subsection{Data preprocessing}\label{subsec4}

The input $X$ feature of the model in this paper is a four-dimensional matrix, respectively derived from physical feature parameters under different penetration working conditions, including warhead type (unit is $kg$), penetration velocity (unit is $m/s$), target material type, and number of penetration layers; the output target $Y$ is the penetration acceleration response feature value under the corresponding working condition, namely the acceleration peak (unit is $g$) and pulse width (unit is $ms$) of each layer. Since there are significant differences in dimensions, numerical ranges, and variation scales between input features and output features, if they are directly used for network training, it is easy to induce problems such as gradient scale imbalance, convergence instability, and a decrease in training efficiency. Therefore, before model training, scale normalization processing is performed on the two types of data, input and output, respectively, to reduce the impact of dimensional differences on the model optimization process and improve the numerical stability of training~\cite{De2023}.

\subsubsection{Normalization of input feature $\mathbf{X}$}\label{subsubsec4}

For the input feature $\mathbf{X} = [x_1, x_2, x_3, x_4]$, this paper performs linear normalization on each feature individually, compressing features of different dimensions into the $[0,1]$ interval to eliminate the influence of numerical scale on model training. The formula for linear normalization is shown as follows:

\begin{equation}
    x'_i = \frac{x_i - x_{\min}}{x_{\max} - x_{\min}}.
\end{equation}
Wherein, $x_i$ is the original feature value, $x_{\min}$ and $x_{\max}$ are the maximum and minimum values of this feature value in the overall sample of this feature, respectively, and the normalized $x'_i \in [0,1]$.

\subsubsection{Normalization of output target $\mathbf{Y}$}\label{subsubsec4}

The output targets $\mathbf{Y} = [y_{acc}, y_{width}]$ represent the acceleration peak and pulse width, respectively. Since the acceleration peak has a large span under different working conditions, reaching a magnitude of tens of thousands, which easily leads to gradient explosion or numerical oscillation during model training, this paper adopts logarithmic normalization for compression, and its formula is shown as follows:

\begin{equation}
    y'_{acc} = \frac{\ln(1 + y_{acc})}{\max(\ln(1 + y_{acc}))}.
\end{equation}

This processing method can effectively suppress the influence of extreme value samples on model training and stabilize gradient changes. For the pulse width feature, its value range is small and non-negative, with the minimum value approaching 0; therefore, the max normalization commonly used for time is adopted, and its formula is:

\begin{equation}
    y'_{width} = \frac{y_{width}}{\max(y_{width})}.
\end{equation}

This normalization method simplifies the calculation process while maintaining the proportional consistency between time scales, allowing time features under different working conditions to be modeled under a unified scale.

\subsubsection{Preservation of normalization parameters and denormalization}
\label{subsubsec4}

To ensure that the output of the model in the inference phase can be accurately mapped back to the real physical quantity scale, all scale parameters involved in the normalization links (including the minimum and maximum values of each input feature and the logarithmic transformation reference values of the acceleration peak feature) are recorded during the training process and saved in the parameter file in the same directory as the model. The inference phase strictly performs reverse scale mapping based on these recorded scale parameters, thereby guaranteeing the consistency of the normalization and denormalization processes and avoiding prediction bias caused by scale mismatch. The input of the final model is $\mathbf{X}'$ and the output is $\mathbf{Y}'$; the model fits the nonlinear function mapping between input physical features and acceleration responses through the deep learning mechanism:

\begin{equation}
    \mathrm{f}_{\theta}: \mathbf{X}' \to \mathbf{Y}'.
\end{equation}

That is, by inputting the physical feature parameters of the warhead and the target, the fast prediction of key features such as the acceleration peak and pulse width of each penetration layer under the corresponding working condition is realized.

\subsection{Performance Evaluation Metrics}
\label{subsec4}

To comprehensively evaluate the performance of the proposed model in the acceleration feature prediction task, this paper quantitatively analyzes the model prediction results from the three dimensions of error amplitude, relative error, and goodness of fit. Comprehensively considering the physical meanings and applicable scenarios of different metrics, the mainly adopted performance evaluation metrics include the Mean Absolute Percentage Error ($MAPE$), Root Mean Squared Error ($RMSE$), and the coefficient of determination $R^2$. $MAPE$ is used to reflect the average relative deviation of prediction results relative to real values, and is applicable to the error comparison between data of different dimensions; $RMSE$ reflects the root mean squared amplitude of the overall prediction error, and is more sensitive to outliers; while $R^2$ is used to measure the explanatory ability of the model for the total variance, serving as an important indicator for comprehensively evaluating the prediction fitting degree, and its calculation formula is shown as follows~\cite{Chicco2021}:

\begin{equation}
    MAPE = \frac{1}{N} \sum_{i=1}^{N} \left| \frac{\hat{y}_i - y_i}{y_i} \right| \times 100\%.
\end{equation}

\begin{equation}
    RMSE = \sqrt{\frac{1}{N} \sum_{i=1}^{N} (\hat{y}_i - y_i)^2}.
\end{equation}

\begin{equation}
    R^2 = 1 - \frac{\sum_{i=1}^{N} (\hat{y}_i - y_i)^2}{\sum_{i=1}^{N} (y_i - \bar{y})^2}.
\end{equation}
Wherein, $\hat{y}_i$ and $y_i$ are the predicted value and actual value of the $i$-th sample respectively, $\bar{y}$ is the mean of the sample true values, and $N$ is the total number of test samples. The smaller the values of $MAPE$ and $RMSE$, the lower the deviation between the predicted value and the true value, and the higher the model accuracy; while the closer $R^2$ is to 1, the better the fitting degree of the model to the actual data.

Furthermore, to eliminate the influence caused by the inconsistency of error scales between different dimensions and different physical quantities, this paper further calculates the Normalized Root Mean Squared Error ($NRMSE$), and its calculation formula is as follows:

\begin{equation}
    NRMSE = \frac{RMSE}{y_{\max} - y_{\min}}.
\end{equation}
Wherein, $y_{\max}$ and $y_{\min}$ are the maximum and minimum values of the target values in the real samples, respectively. $NRMSE$ reflects the error level of the model under a unified scale, which helps to conduct performance comparisons between different prediction objects.

Integrating the above metrics, this paper systematically evaluates the performance of the model from three aspects: error precision, result stability, and model fitting ability, and ensures the reliability and generalization of the results through multi-fold cross-validation.

\subsection{Model Training}
\label{subsec4}

Based on the PyTorch deep learning framework, this paper constructs a three-layer SE-MLP model, which is used for the nonlinear mapping learning between physical parameters under different working conditions and acceleration features. The model input is the physical feature matrix after unified scale processing, and the output is the acceleration peak and pulse width corresponding to each penetration layer, realizing the nonlinear mapping relationship from working condition physical parameters to acceleration feature values.

In the training phase, all samples are subjected to fold processing, and a four-fold cross-validation ($K=4$) strategy is adopted to ensure the generalization performance and stability of the model. In each fold verification, 75\% of the samples are randomly partitioned as the training set, and 25\% of the samples act as the validation set, ensuring that the model can achieve robust learning under multi-working condition and multi-layer target conditions.

The data Batch Size is set to 32, and the maximum training epochs are 200. The model adopts the AdamW optimizer for parameter updating, the initial learning rate is set to $1 \times 10^{-3}$, the weight decay coefficient is $1 \times 10^{-4}$, and the ReduceLROnPlateau learning rate scheduling strategy is introduced; the learning rate decay coefficient is 0.5, and the number of tolerance stagnation epochs is 15, that is, when the validation set loss does not decrease within 15 consecutive epochs, the learning rate is automatically halved to improve the convergence speed and stability of training. To avoid parameter update stagnation caused by excessive decay of the learning rate, the learning rate lower limit threshold is set to $1 \times 10^{-6}$.

The loss function adopts Weighted Mean Squared Error ($WMSE$) to simultaneously account for the precision requirements of the two types of prediction targets: acceleration peak and pulse width. Considering that the acceleration peak has higher importance in penetration layer counting, interlayer identification, and burst point control strategies, and its magnitude is obviously larger than the pulse width feature, loss weights of 0.7 and 0.3 are assigned to the acceleration peak and pulse width respectively, thereby enhancing the model's sensitivity to the acceleration peak. The final loss function can be expressed as:

\begin{equation}
    L = 0.7 \times MSE_{\text{acc}} + 0.3 \times MSE_{\text{width}}.
\end{equation}

To further enhance training stability and reduce the risk of overfitting, Batch Normalization and Dropout (dropout rate is set to 0.1) are added to each fully connected layer of the model, effectively promoting gradient flow and suppressing feature overfitting. GELU is selected as the activation function, possessing both the sparsity of ReLU and the smoothness of Sigmoid, which can improve the nonlinear expression ability of the model while maintaining gradient stability. During the training process, the variation trends of training loss and validation loss are monitored in real time to prevent overfitting, and the best model weights are recorded for final testing and performance evaluation.

To quantitatively evaluate the prediction ability of the constructed SE-MLP model under different working conditions, Fig. 7 and Table 3 respectively present the performance metric results of the 4-fold cross-validation and their visual representations. The table summarizes key evaluation metrics such as the Mean Absolute Percentage Error ($MAPE$), Root Mean Squared Error ($RMSE$), coefficient of determination ($R^2$), and Normalized Root Mean Squared Error ($NRMSE$) of the model in each fold of verification.

\begin{figure}[t]
    \centering
    
    \begin{subfigure}[b]{0.48\textwidth}
        \centering
        \includegraphics[width=\linewidth]{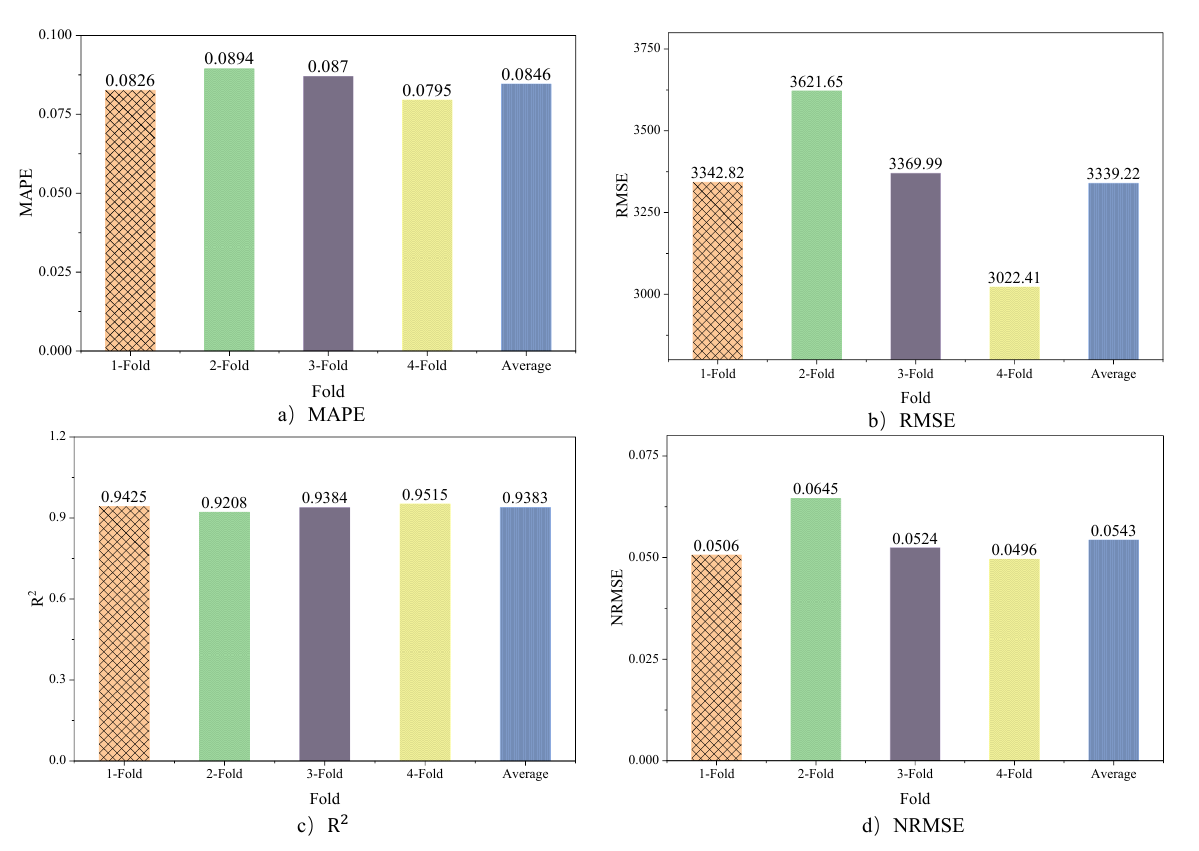}
        \caption{Comparison of evaluation metrics for acceleration peak.}
        \label{fig:peak_metrics}
    \end{subfigure}
    \hfill 
    \begin{subfigure}[b]{0.48\textwidth}
        \centering
        \includegraphics[width=\linewidth]{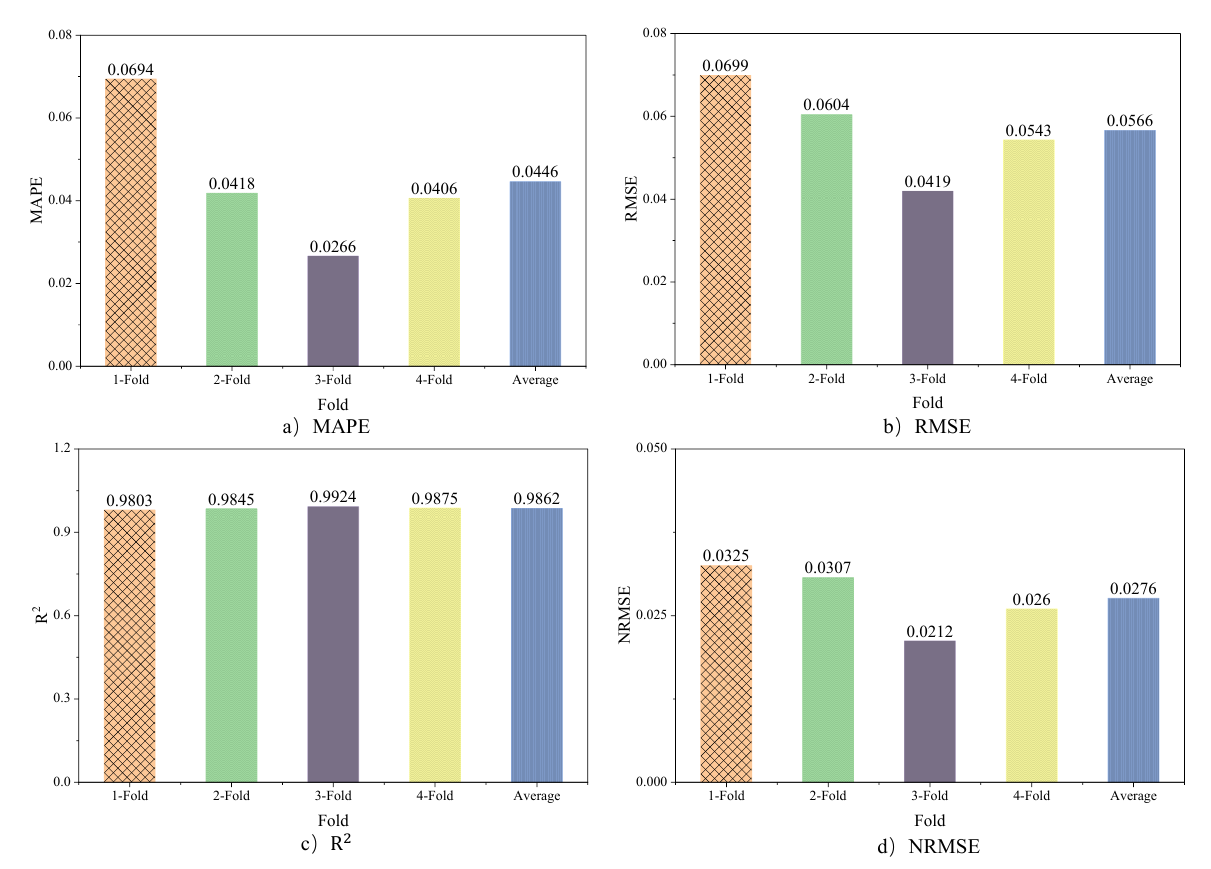}
        \caption{Comparison of evaluation metrics for pulse width.}
        \label{fig:pulse_metrics}
    \end{subfigure}
    
    \caption{Comparison of key evaluation metrics for penetration features.}
    \label{fig:metrics_comparison}
\end{figure}

It can be seen from the results in the table that the prediction performance of the model in the four-fold cross-validation is stable, and the fluctuations of various metrics are small, reflecting good generalization. Wherein, the average $R^2$ of acceleration peak prediction reaches 0.9383, indicating that the model can accurately fit the nonlinear mapping relationship between physical parameters and acceleration features; its average $MAPE$ is 8.46\%, and $NRMSE$ is 0.0543, demonstrating that the overall error of the model is controlled within a reasonable range. In contrast, the prediction performance of pulse width is more excellent, with an average $R^2$ as high as 0.9862 and a $MAPE$ of only 4.46\%, indicating that the model possesses higher sensitivity and robustness in time-domain feature modeling.

Furthermore, it can be seen from the comparison of the four-fold results that the model can maintain stable convergence under different working conditions, and the prediction error does not show obvious abnormal fluctuations, indicating that the constructed three-layer SE-MLP model possesses good robustness and adaptability. In summary, the model exhibits excellent accuracy and stability in the prediction of the two penetration features, acceleration peak and pulse width, providing reliable theoretical and methodological support for the prediction of penetration acceleration feature values based on physical parameters.

\begin{table}[t]
    \centering
    \caption{Comparison of K-Fold evaluation metrics.}
    \label{tab:kfold_metrics}
    
    \begin{tabular*}{\textwidth}{@{\extracolsep{\fill}}lcccccccc}
        \toprule
        \multirow{2}{*}{Fold} & \multicolumn{4}{c}{Acceleration peak} & \multicolumn{4}{c}{Pulse width} \\
        
        \cmidrule(lr){2-5} \cmidrule(lr){6-9}
        
         & $MAPE$ & $RMSE$ & $R^2$ & $NRMSE$ & $MAPE$ & $RMSE$ & $R^2$ & $NRMSE$ \\
        \midrule
        
        1-Fold & 8.26\% & 3342.82 & 0.9425 & 0.0506 & 6.94\% & 0.0699 & 0.9803 & 0.0325 \\
        2-Fold & 8.94\% & 3621.65 & 0.9208 & 0.0645 & 4.18\% & 0.0604 & 0.9845 & 0.0307 \\
        3-Fold & 8.70\% & 3369.99 & 0.9384 & 0.0524 & 2.66\% & 0.0419 & 0.9924 & 0.0212 \\
        4-Fold & 7.95\% & 3022.41 & 0.9515 & 0.0496 & 4.06\% & 0.0543 & 0.9875 & 0.0260 \\
        
        \textbf{Average} & \textbf{8.46\%} & \textbf{3339.22} & \textbf{0.9383} & \textbf{0.0543} & \textbf{4.46\%} & \textbf{0.0566} & \textbf{0.9862} & \textbf{0.0276} \\
        \bottomrule
    \end{tabular*}
\end{table}

\subsection{Model Performance Comparison and Ablation Experiments}
\label{subsec1}
To verify the robustness and adaptability of the proposed SE-MLP model in the penetration acceleration feature prediction task, this paper systematically analyzes and compares from two dimensions: horizontal model comparison and longitudinal ablation experiments. The former aims to evaluate the comprehensive performance of SE-MLP compared with other mainstream neural network models in terms of prediction accuracy and stability; the latter is used to investigate the contribution degree of the SE channel attention module and residual structure to the overall prediction performance.

\subsubsection{Model performance comparison}
\label{subsec1}
To comprehensively evaluate the predictive performance and generalization capability of the proposed model, Transformer, XGBoost, Random Forest (RF), and Support Vector Regression (SVR) were selected as benchmarking baselines. The Transformer model possesses powerful global dependency modeling ability and can capture long-range correlations between input features through the self-attention mechanism, but its parameter scale is large, it relies significantly on data volume and computational resources, and it is prone to overfitting in the case of limited sample size~\cite{Xu2024,ChenL2022}. In contrast, the XG-Boost model has better stability and interpretability under small sample conditions, but its tree-based piecewise regression characteristics limit the deep nonlinear characterization ability for continuous feature spaces~\cite{Cheng2024,Khan2023}. RF effectively mitigates the risk of overfitting inherent in single models by integrating multiple decision trees, generally demonstrating robustness on small-to-medium datasets. However, its ability to model complex feature couplings relies heavily on tree depth, making it difficult to fully capture high-dimensional continuous nonlinear mappings. SVR, based on the principle of structural risk minimization, exhibits stability in high-dimensional, small-sample regression tasks. Yet, its performance is highly sensitive to the choice of kernel functions and hyperparameters. In scenarios involving high feature dimensionality and strong nonlinearity, parameter tuning becomes arduous, often leading to performance bottlenecks~\cite{Feng2020,Bargam2024}. Existing methods for acquiring penetration overload characteristics predominantly rely on numerical simulations or empirical modeling, with relatively few systematic studies based on deep learning. Consequently, building upon a review and reproduction of traditional machine learning methods, this study implements representative models—including Transformer, XGBoost, RF, and SVR—as benchmarks. This comparative framework aims to systematically evaluate the adaptability, generalization capability, and comprehensive performance advantages of the proposed SE-MLP model in penetration overload prediction tasks.

\begin{figure}[t]
    \centering
    
    \begin{subfigure}[b]{0.48\textwidth}
        \centering
        \includegraphics[width=\linewidth]{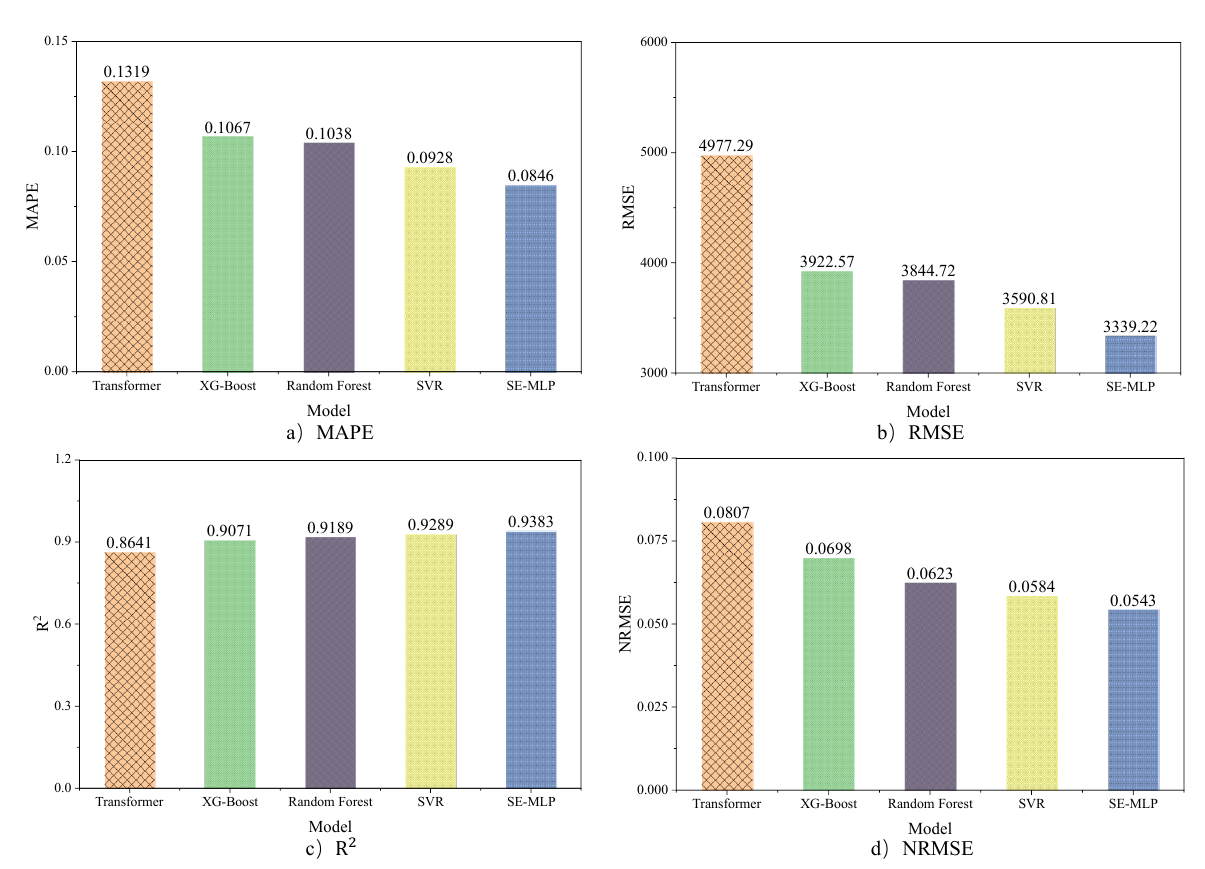}
        \caption{Comparison of evaluation metrics for acceleration peak.}
        \label{fig:peak_comparison_models}
    \end{subfigure}
    \hfill 
    \begin{subfigure}[b]{0.48\textwidth}
        \centering
        \includegraphics[width=\linewidth]{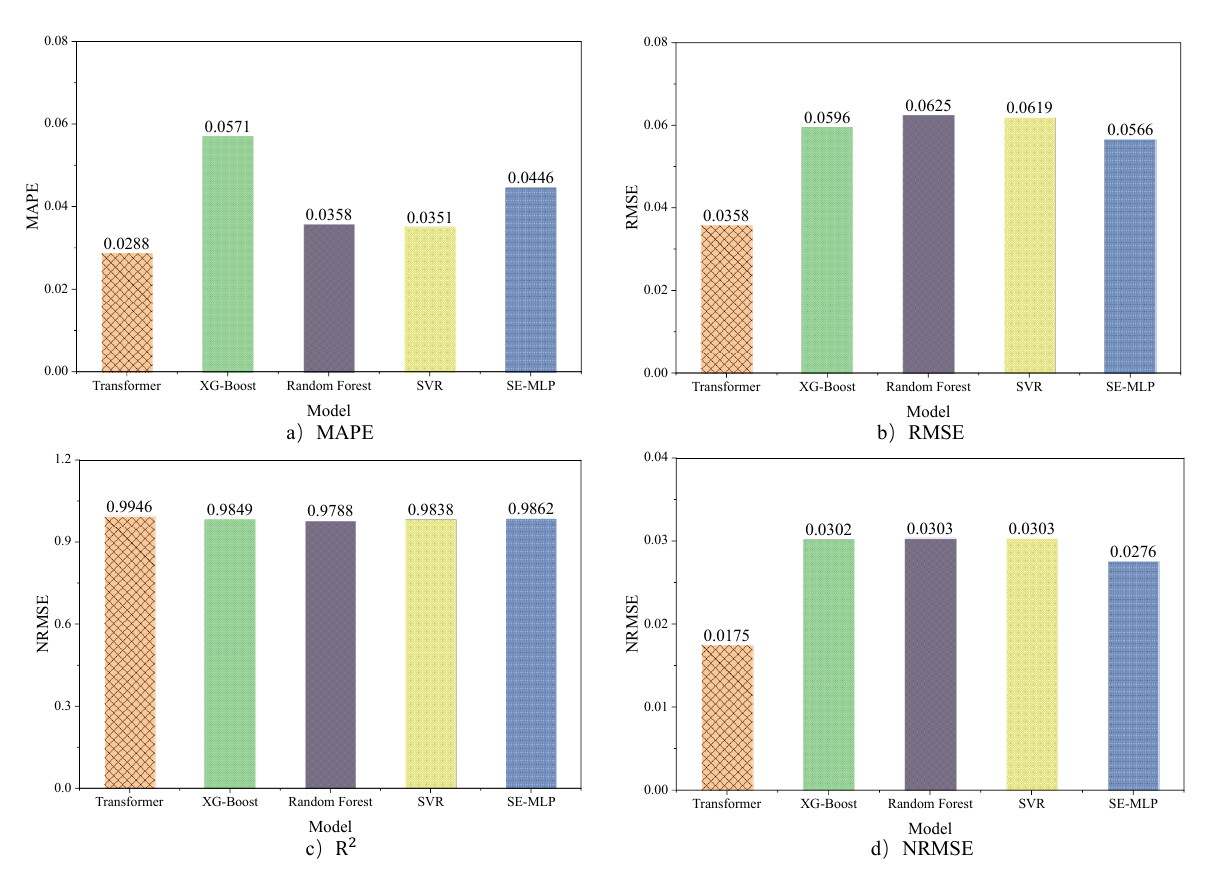}
        \caption{Comparison of evaluation metrics for pulse width.}
        \label{fig:pulse_comparison_models}
    \end{subfigure}
    
    \caption{Comparison of key evaluation metrics for penetration features of different models.}
    \label{fig:model_comparison}
\end{figure}

Under the same dataset, training parameters, and evaluation metrics, comparative experiments were conducted on Transformer, XG-Boost, and the proposed SE-MLP model. The experimental conditions remain consistent: the batch size is 32, the maximum training epochs are 200, the optimizer is AdamW, the initial learning rate is set to $1 \times 10^{-3}$, and four-fold cross-validation is adopted to obtain robust average results. All evaluation metrics are calculated based on the prediction results after denormalization. The comparison results of model evaluation metrics are shown in Fig. 8 and Table 4, respectively.

\begin{table}[t]
    \centering

    \small 
    \setlength{\tabcolsep}{2pt}
    
    \caption{Comparison of evaluation metrics of different models.}
    \label{tab:model_comparison_metrics}

    \begin{tabular*}{\textwidth}{@{\extracolsep{\fill}}lcccccccc}
        \toprule
        \multirow{2}{*}{Model Type} & \multicolumn{4}{c}{Acceleration peak} & \multicolumn{4}{c}{Pulse width} \\
        \cmidrule(lr){2-5} \cmidrule(lr){6-9}
        
        & $MAPE$ & $RMSE$ & $R^2$ & $NRMSE$ & $MAPE$ & $RMSE$ & $R^2$ & $NRMSE$ \\
        \midrule
        
        Transformer~\cite{Xu2024,ChenL2022}   & 13.19\% & 4977.29 & 0.8641 & 0.0807 & 2.88\% & 0.0358 & 0.9946 & 0.0175 \\
        XG-Boost~\cite{Cheng2024,Khan2023}    & 10.67\% & 3922.57 & 0.9071 & 0.0698 & 5.71\% & 0.0596 & 0.9849 & 0.0302 \\
        Random Forest~\cite{Feng2020}         & 10.38\% & 3844.72 & 0.9189 & 0.0623 & 3.58\% & 0.0625 & 0.9788 & 0.0303 \\
        SVR~\cite{Bargam2024}                 & 9.28\%  & 3590.81 & 0.9289 & 0.0584 & 3.51\% & 0.0619 & 0.9838 & 0.0303 \\
        
        \textbf{SE-MLP} & \textbf{8.46\%} & \textbf{3339.22} & \textbf{0.9383} & \textbf{0.0543} & \textbf{4.46\%} & \textbf{0.0566} & \textbf{0.9862} & \textbf{0.0276} \\
        \bottomrule
    \end{tabular*}
\end{table}

As shown in Table 4, all five models achieved relatively high accuracy in the prediction tasks for both acceleration peak and pulse width; however, significant differences in overall performance exist. A comprehensive comparison reveals that the SE-MLP model exhibits optimal performance in both prediction tasks, demonstrating its distinct advantages in predicting acceleration signal feature values.

In terms of acceleration peak prediction, the SE-MLP model achieved an $RMSE$ of 3339.22, a $MAPE$ of 8.46\%, and an $R^2$ of 0.9383, outperforming all comparative models. This indicates that the SE-MLP can more accurately fit high-amplitude impact features under different working conditions. Its normalized error, $NRMSE$, is merely 0.0543, further indicating a minimal relative deviation between the predicted results and the ground truth, ensuring high numerical consistency. These results reflect that the SE-MLP model possesses stronger feature extraction and robust fitting capabilities when processing complex nonlinear inputs and multi-scale feature variations.

Regarding pulse width prediction, the SE-MLP model also exhibited excellent performance, with an $R^2$ reaching 0.9862 and $MAPE$ controlled at 4.46\%. The overall prediction accuracy improved by approximately 11.7\% compared with the standard MLP, and although marginally lower than the Transformer by 1.6\%, the fluctuation range is smaller, offering superior stability. Specifically, regarding the $RMSE$ metric, the SE-MLP (0.0566) is slightly higher than the Transformer (0.0358); however, considering the Transformer's instability and pronounced tendency to overfit in peak prediction, the SE-MLP offers greater advantages in comprehensive performance and robustness. The Transformer achieves an $R^2$ as high as 0.9946 in pulse width prediction, benefitting largely from the adaptability of its powerful self-attention mechanism to time-domain features. Yet, in modeling amplitude responses such as acceleration peak, limited input samples and sensitivity to learning rates cause its global feature modeling to introduce overfitting, resulting in substantial performance volatility.

Further analysis indicates that the core reason for the performance improvement of the SE-MLP model lies in the feature selection and gradient propagation optimization brought by structural improvements. The SE module weights input features through the channel attention mechanism, which can adaptively adjust the response amplitudes of different feature channels, making the model focus more on physical parameters highly correlated with output features, such as penetration velocity and layer number, thereby enhancing the contribution of key features and suppressing the interference of redundant information. Furthermore, the introduction of the residual structure effectively mitigates the gradient vanishing problem in deep networks, enabling low-level features to directly participate in high-level decision-making, improving the stability of the training process and model convergence speed. In contrast, traditional MLP is prone to feature degradation phenomena in multi-layer mapping, while Transformer, although capable of capturing global dependencies, easily leads to parameter oscillation and training instability under small sample and high noise conditions.

In summary, the SE-MLP model achieves a comprehensive improvement in prediction accuracy, training stability, and generalization ability under the premise of controllable computational complexity. Its precise prediction of acceleration peak and pulse width features verifies the effectiveness of the channel attention and residual fusion structure in the problem of acceleration signal feature value prediction, providing a high-precision and lightweight solution for penetration acceleration feature value prediction based on working condition parameters.

\subsubsection{Ablation experiments}
\label{subsec1}
To further investigate the influence of key structures in the SE-MLP model on the overall prediction performance, this paper designs three groups of ablation experiments to quantitatively analyze the roles of the SE channel attention module and the residual connection structure, respectively. All experiments are based on the same dataset, training parameters, and evaluation metric settings to guarantee the comparability of results and the reliability of conclusions.
Specifically, the experiment includes the following three network structures:
Three-layer MLP: Only retains the basic fully connected structure without any improvement modules;
Three-layer MLP + SE: Introduces the SE channel attention module after each fully connected layer network, but does not use residual connections;
SE-MLP: Further adds residual connections on the basis of the previous structure, which is the complete model proposed in this paper.
Fig. 9 and Table 5 present the average performance metrics of the three structures in acceleration peak and pulse width prediction. The results show that as the network structure is gradually enhanced, the prediction accuracy and robustness of the model are significantly improved.

\begin{figure}[t]
    \centering
    
    \begin{subfigure}[b]{0.48\textwidth}
        \centering
        \includegraphics[width=\linewidth]{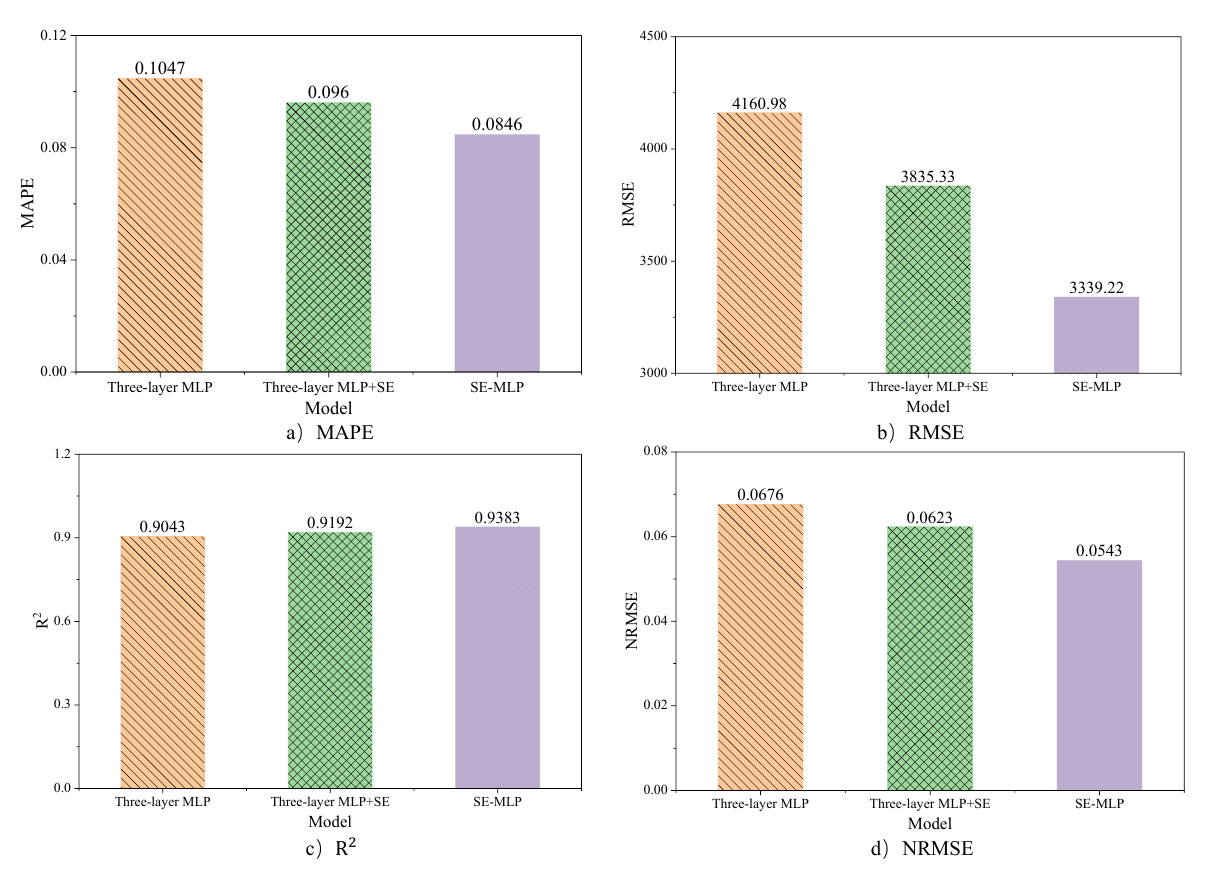}
        \caption{Comparison of evaluation metrics for acceleration peak.}
        \label{fig:ablation_peak}
    \end{subfigure}
    \hfill 
    \begin{subfigure}[b]{0.48\textwidth}
        \centering
        \includegraphics[width=\linewidth]{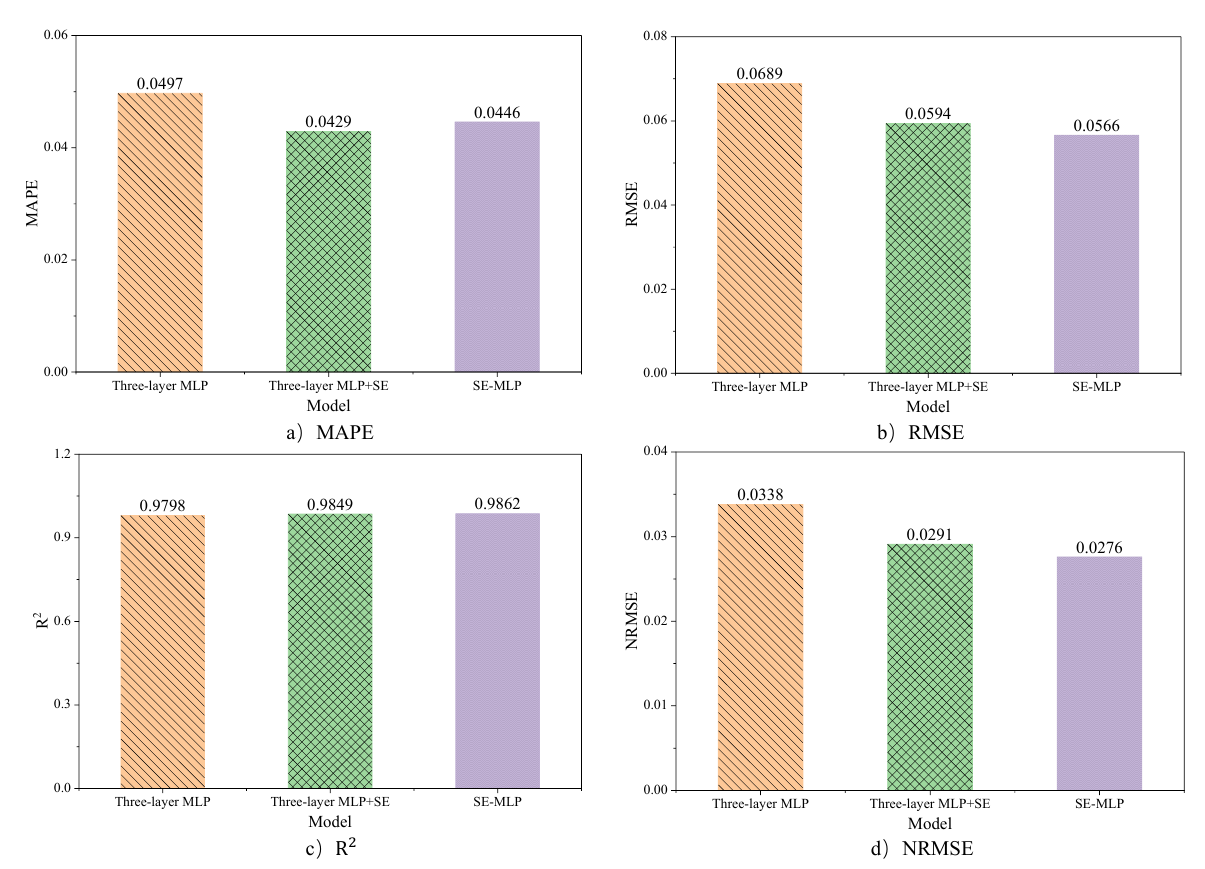}
        \caption{Comparison of evaluation metrics for pulse width.}
        \label{fig:ablation_pulse}
    \end{subfigure}
    
    \caption{Comparison of key evaluation metrics for penetration features in ablation experiments.}
    \label{fig:ablation_comparison}
\end{figure}

It can be seen from the results in the table that with the gradual improvement of the network structure, the overall performance of the model in the acceleration peak and pulse width prediction tasks presents a significant increasing trend. The basic three-layer MLP model has an $RMSE$ of 4160.98 and an $R^2$ of 0.9043 in peak prediction, indicating that although it possesses certain nonlinear mapping ability, the fitting is insufficient; in pulse width prediction, $R^2$ reaches 0.9798, showing certain stability, but there is still large room for improvement in overall accuracy.

After introducing the SE module, the model performance improves obviously. The SE module adaptively adjusts the weight distribution of each input feature through the channel attention mechanism, enabling the model to focus on key features highly correlated with output variables (such as penetration velocity, target material, etc.), thereby effectively suppressing the interference of redundant features on training. Experimental results indicate that the peak prediction $RMSE$ of the MLP+SE model decreases to 3835.33 (reduced by about 7.8\% compared with basic MLP), and $R^2$ increases to 0.9192; the $MAPE$ of pulse width prediction drops to 4.29\%, indicating that the introduction of the SE module significantly enhances the feature selection ability and nonlinear fitting accuracy of the model.

\begin{table}[t]
    \centering
    \small
    
    \setlength{\tabcolsep}{2pt}
    
    \caption{Comparison of evaluation metrics in ablation experiments.}
    \label{tab:ablation_metrics}
    
    \begin{tabular*}{\textwidth}{@{\extracolsep{\fill}}lcccccccc}
        \toprule
        \multirow{2}{*}{Model Type} & \multicolumn{4}{c}{Acceleration peak} & \multicolumn{4}{c}{Pulse width} \\
        \cmidrule(lr){2-5} \cmidrule(lr){6-9}
        
         & $MAPE$ & $RMSE$ & $R^2$ & $NRMSE$ & $MAPE$ & $RMSE$ & $R^2$ & $NRMSE$ \\
        \midrule
        
        Three-layer MLP & 10.47\% & 4160.98 & 0.9043 & 0.0676 & 4.97\% & 0.0689 & 0.9798 & 0.0338 \\
        Three-layer MLP + SE & 9.60\% & 3835.33 & 0.9192 & 0.0623 & 4.29\% & 0.0594 & 0.9849 & 0.0291 \\
        
        \textbf{SE-MLP} & \textbf{8.46\%} & \textbf{3339.22} & \textbf{0.9383} & \textbf{0.0543} & \textbf{4.46\%} & \textbf{0.0566} & \textbf{0.9862} & \textbf{0.0276} \\
        \bottomrule
    \end{tabular*}
\end{table}

After further adding the residual connection structure on this basis, the model performance obtains continuous improvement. The residual path effectively mitigates the problems of gradient vanishing and feature degradation in deep networks by establishing direct connections between low-level and high-level features, making the training process more stable and the convergence speed faster. The peak prediction $RMSE$ of the complete SE-MLP model further decreases to 3339.22 (reduced by 12.9\% compared with MLP+SE, and reduced by 19.7\% compared with basic MLP), and $R^2$ increases to 0.9383; the $RMSE$ and $NRMSE$ of pulse width prediction drop to 0.0566 and 0.0276 respectively, exhibiting superior accuracy and robustness. This result indicates that the residual structure plays a key role in improving the generalization ability and training stability of the model, making the learning of high-dimensional input features of the model under complex penetration environments more sufficient.

Therefore, the SE channel attention and the residual structure play a synergistic role in penetration acceleration feature prediction, which is the important structural basis for the SE-MLP model performance being significantly superior to the traditional fully connected network. This structural design idea provides new theoretical support and implementation paths for the prediction of penetration acceleration feature values.

\section{Simulation and Experimental Verification of Prior Feature Values}

To further verify the applicability and prediction accuracy of the SE-MLP model, additional numerical simulations and range recovery test data were used to evaluate the trained model. Specifically, the prediction accuracy was assessed by comparing the predicted acceleration peaks and pulse widths with the measured data from both the simulations and the range tests.
To ensure consistency and comparability, the simulation parameters were strictly aligned with the range recovery test conditions, thereby verifying the model's generalization ability in a real penetration environment. The specific test parameters were: a 94 kg-class warhead penetrating a 5-layer C40 concrete target at an initial velocity of 926 m/s. The first target layer had a thickness of 0.3 m, while the remaining four layers were 0.18 m each. The comparison results are presented in Fig. 10 and Table 6.

\begin{figure}[t]
    \centering
    \begin{subfigure}{0.48\textwidth}
        \centering
        \includegraphics[width=\textwidth]{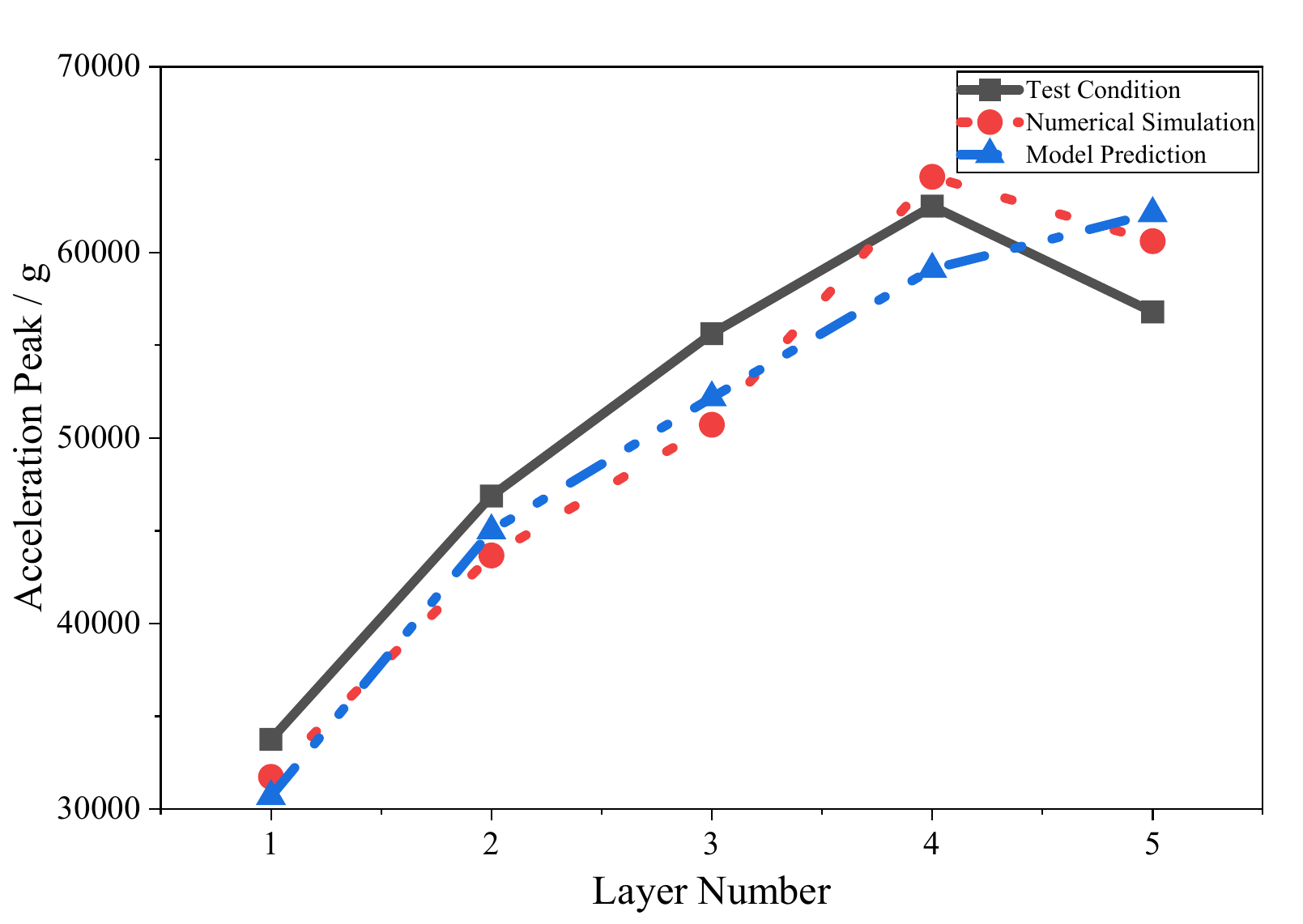}
        \caption{Comparison diagram of acceleration peak test.}
        \label{fig:test_peak}
    \end{subfigure}
    \hfill 
    \begin{subfigure}{0.48\textwidth}
        \centering
        \includegraphics[width=\textwidth]{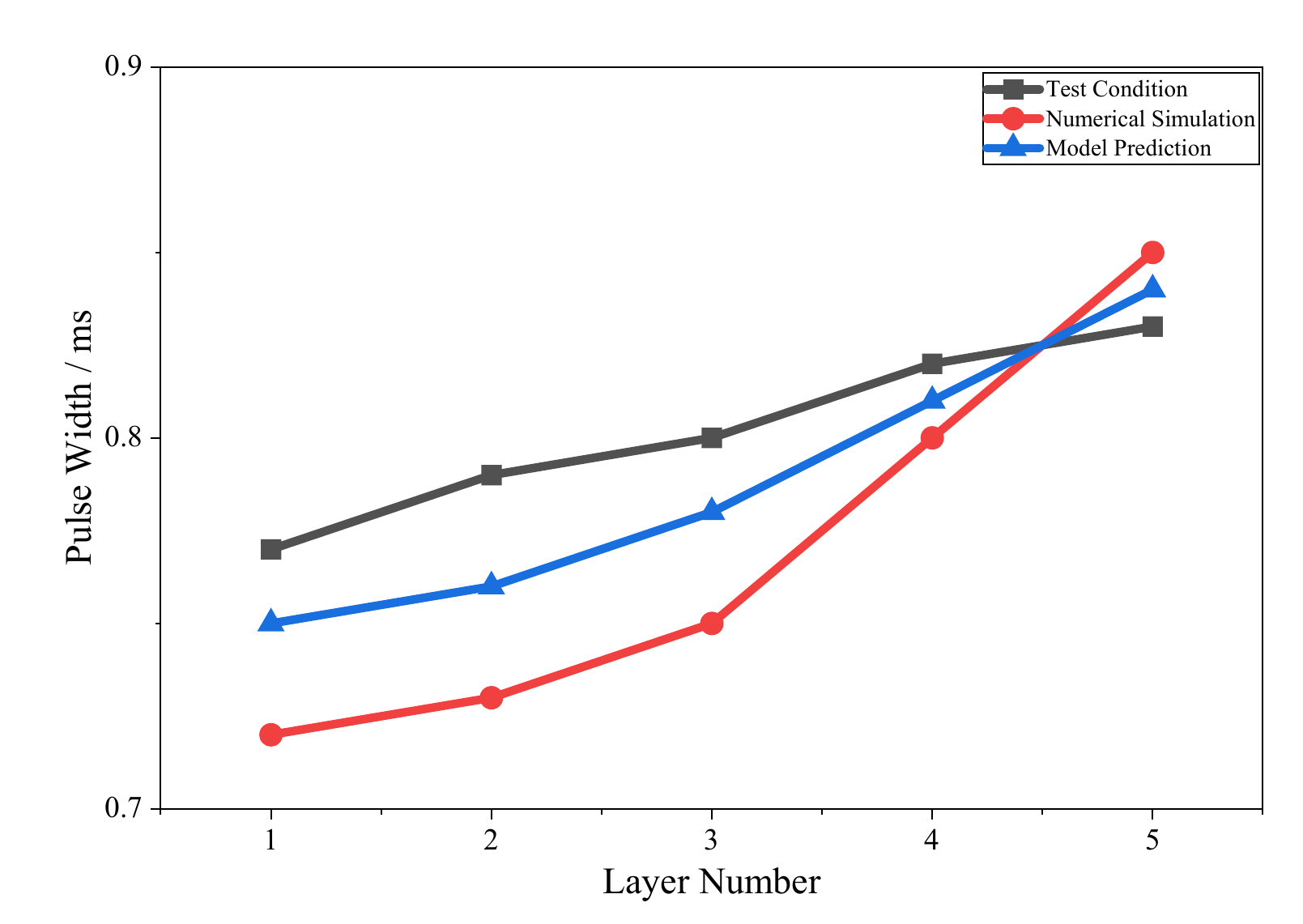}
        \caption{Comparison diagram of pulse width test.}
        \label{fig:test_pulse}
    \end{subfigure}
    
    \caption{Test comparison.}
    \label{fig:test_comparison}
\end{figure}

\begin{table}[t]
    \centering
    \caption{Test comparison results and error table.}
    \label{tab:test_results}
    
    \begin{tabular*}{\textwidth}{@{\extracolsep{\fill}}lcccccc}
        \toprule
        \makecell{Acceleration \\ feature} & \makecell{Layer \\ number} & \makecell{Test \\ condition} & \makecell{Numerical \\ simulation} & \makecell{Model \\ prediction} & \makecell{Simulation \\ error} & \makecell{Prediction \\ error} \\
        \midrule
        
        \multirow{5}{*}{\makecell{Acceleration \\ peak/g}} 
        & 1st layer & 33750 & 31734 & 30682 & \textbf{5.97\%} & \textbf{9.09\%} \\
        & 2nd layer & 46875 & 43673 & 42006 & \textbf{6.83\%} & \textbf{10.39\%} \\
        & 3rd layer & 55625 & 50714 & 51178 & \textbf{8.83\%} & \textbf{7.99\%} \\
        & 4th layer & 62500 & 64081 & 58102 & \textbf{2.53\%} & \textbf{7.04\%} \\
        & 5th layer & 56785 & 60612 & 64707 & \textbf{6.57\%} & \textbf{13.95\%} \\
        
        \midrule
        
        \multirow{5}{*}{\makecell{Pulse \\ width/ms}} 
        & 1st layer & 0.77 & 0.72 & 0.75 & \textbf{6.49\%} & \textbf{2.60\%} \\
        & 2nd layer & 0.79 & 0.73 & 0.76 & \textbf{7.59\%} & \textbf{3.80\%} \\
        & 3rd layer & 0.80 & 0.75 & 0.78 & \textbf{6.25\%} & \textbf{2.50\%} \\
        & 4th layer & 0.82 & 0.80 & 0.81 & \textbf{2.44\%} & \textbf{1.22\%} \\
        & 5th layer & 0.83 & 0.85 & 0.84 & \textbf{2.41\%} & \textbf{1.20\%} \\
        \bottomrule
    \end{tabular*}
\end{table}

It can be seen from Fig. 10 and Table 6 that the prediction results of the SE-MLP model for acceleration peak and pulse width are highly consistent with the experimental measured values, the peak error is controlled within 15\%, and the pulse width errors are all less than 4\%. The model can accurately capture the variation laws between different layers, reflecting good physical consistency and prediction stability. The results show that the constructed SE-MLP model has strong generalization ability and engineering applicability under actual penetration working conditions, and can provide reliable support for the acquisition of penetration prior feature values.

\section{Conclusion}

To address the challenges of long simulation cycles and high computational costs associated with acquiring penetration prior feature values, this paper proposes a lightweight multi-layer perceptron (SE-MLP) prediction model incorporating channel attention and residual fusion. By synergizing the nonlinear mapping capability of MLPs with the adaptive feature weighting of the SE mechanism, the model achieves an efficient mapping between physical parameters and penetration acceleration  features. The main conclusions are as follows:

\begin{itemize}
    \item[\textcolor{black}{$\bullet$}] 
    
The proposed SE-MLP integrates SE attention mechanisms with residual structures to facilitate adaptive feature weighting and cross-layer information transfer. This architecture effectively enhances feature extraction capabilities while ensuring training stability.

    \item[\textcolor{black}{$\bullet$}] 
    
Benchmarking against MLP, Transformer, and XGBoost demonstrates that SE-MLP achieves superior performance in predicting penetration overload characteristics. It reduced the average $RMSE$ by 15\%–30\% and improved $R^2$ by approximately 4\%, highlighting its exceptional accuracy and generalization.

    \item[\textcolor{black}{$\bullet$}] 
    
Ablation studies confirmed the synergistic effect between channel attention and residual structures. Removing either component resulted in an $RMSE$ increase of 10\%–15\% and an $R^2$ decrease of 2\%–4\%, indicating their critical role in maintaining predictive precision.

    \item[\textcolor{black}{$\bullet$}] 
    
Validation via numerical simulations and field tests confirms that prediction errors remain within 15\%. The model accurately captures energy transfer dynamics and interlayer feature variations during the penetration process.

\end{itemize}

Despite the promising prediction performance, there is potential for further improvement. Future research will primarily focus on expanding the physical feature space by incorporating multidimensional parameters—such as attitude angle, length-to-diameter ratio, and frequency domain response—to construct a framework with richer physical significance.

\bibliography{sn-bibliography}      
\end{document}